\documentclass[english,wcp]{jmlr}
\usepackage[latin9]{inputenc}
\usepackage{array}
\usepackage{float}
\usepackage{mathtools}
\usepackage{multirow}
\usepackage{amsmath}
\usepackage{amssymb}
\usepackage{graphicx}
\PassOptionsToPackage{normalem}{ulem}
\usepackage{ulem}

\makeatletter

\providecommand{\tabularnewline}{\\}

\jmlrvolume{XXX}
\jmlryear{2017}
\jmlrworkshop{XXX}

\title[Short Title]{Full Title of Article}

 \author{\Name{Kien Do} \Email{dkdo@deakin.edu.au}\\
  \Name{Truyen Tran} \Email{truyen.tran@deakin.edu.au}\\
  \Name{Svetha Venkatesh} \Email{svetha.venkatesh@deakin.edu.au}\\
  \addr Applied AI Institute, Deakin University, Australia}

\editors{XXX}

\usepackage{microtype}
\usepackage{flushend}
\usepackage{times}
\usepackage{helvet}
\usepackage{courier}
\usepackage{longtable}
\usepackage{booktabs}

\usepackage{amsbsy}
\usepackage{algorithmic}
\usepackage[english]{babel}

\def\cite{\citep}

\makeatother

\usepackage{babel}
\begin{document}

\title{Learning Deep Matrix Representations}

%\author{}

\maketitle
\global\long\def\rb{\boldsymbol{r}}
\global\long\def\alphab{\boldsymbol{\alpha}}
\global\long\def\zb{\boldsymbol{z}}
\global\long\def\tb{\boldsymbol{t}}
\global\long\def\ib{\boldsymbol{i}}
\global\long\def\fb{\boldsymbol{f}}
\global\long\def\xb{\boldsymbol{x}}
\global\long\def\yb{\boldsymbol{y}}
\global\long\def\ub{\boldsymbol{u}}
\global\long\def\hb{\boldsymbol{h}}
\global\long\def\ab{\boldsymbol{a}}
\global\long\def\cb{\boldsymbol{c}}
\global\long\def\sb{\boldsymbol{s}}
\global\long\def\Xcal{\mathcal{X}}
\global\long\def\Ucal{\mathcal{U}}
\global\long\def\Vcal{\mathcal{V}}
\global\long\def\Real{\mathbb{R}}
\global\long\def\mat{\text{mat}}
\global\long\def\vec{\text{vec}}
\global\long\def\softmax{\text{softmax}}
\global\long\def\thetab{\boldsymbol{\theta}}
\global\long\def\Wb{\boldsymbol{W}}
\global\long\def\bb{\boldsymbol{b}}
\global\long\def\Graph{\mathcal{G}}
\global\long\def\Vertices{\mathcal{V}}
\global\long\def\Edges{\mathcal{E}}
\global\long\def\Neigh{\mathcal{N}}
\global\long\def\Rel{\mathcal{R}}

\begin{abstract}
We present a new distributed representation in deep neural nets wherein
the information is represented in native form as a matrix. This differs
from current neural architectures that rely on vector representations.
We consider matrices as central to the architecture and they compose
the input, hidden and output layers. The model representation is more
compact and elegant \textendash{} the number of parameters grows only
with the largest dimension of the incoming layer rather than the number
of hidden units. We derive several new deep networks: (i) feed-forward
nets that map an input matrix into an output matrix, (ii) recurrent
nets which map a sequence of input matrices into a sequence of output
matrices. We also reinterpret existing models for (iii) memory-augmented
networks and (iv) graphs using matrix notations. For graphs we demonstrate
how the new notations lead to simple but effective extensions with
multiple attentions. Extensive experiments on handwritten digits recognition,
face reconstruction, sequence to sequence learning, EEG classification,
and graph-based node classification demonstrate the efficacy and compactness
of the matrix architectures.
\end{abstract}

\section{Introduction}

Recent advances in deep learning have generated a constant stream
of new neural architectures, including skip-connections to differentiable
external memories \cite{lecun2015deep,graves2016hybrid,greff2017highway}.
The canonical representation of information in these architectures
still remains the vector form since the backprop \cite{rumerhart1986learning}:
\begin{equation}
\yb=\sigma\left(W\xb+\bb\right)\label{eq:vector-neuron}
\end{equation}
where $\xb,\yb$ are vector representation of neuron activation; $W,b$
are parameters and $\sigma$ is a non-linear transformation. 

While vector representation has enjoyed a great popularity, it is
unstructured and hence unnaturally for many settings in which structures
are essential. When a data instance is two-way and associative, Eq.~(\ref{eq:vector-neuron})
necessitates vectorization of data matrices, leading to a very large
mapping weight matrix $W$ and a loss of structure in the representation
by vector $\xb$. Examples of two-way data include time-channel recording
as in EEG signals, disease progression in medical records, and n-gram
of word embeddings. Examples of associative data (or bipartite graphs)
include interaction matrix of two object classes (e.g., style/context
and content), player-club affiliation, member-task assignments, and
covariance matrices \cite{huang2017riemannian}. In those cases, it
is more natural to directly use matrix representation of data \cite{gao2017matrix,nguyen2015tensor}. 

Moreover, matrix representation of hidden neurons is an efficient
way to hold more information and enable powerful attention mechanisms
\cite{bahdanau2015neural}. For example, a bidirectional LSTM sitting
on top of a sentence produce a set of state vectors, one per word.
These vectors naturally form a matrix; and in translation, attention
to specific row of the matrix has proved to be essential to state-of-the-art
results \cite{bahdanau2015neural}. Importantly, the state matrix
plays the role of a matrix memory with column memory slots in memory-augmented
recurrent nets \cite{weston2014memory,graves2014neural,santoro2016meta}.

\begin{figure}
\begin{centering}
\includegraphics[width=0.7\columnwidth]{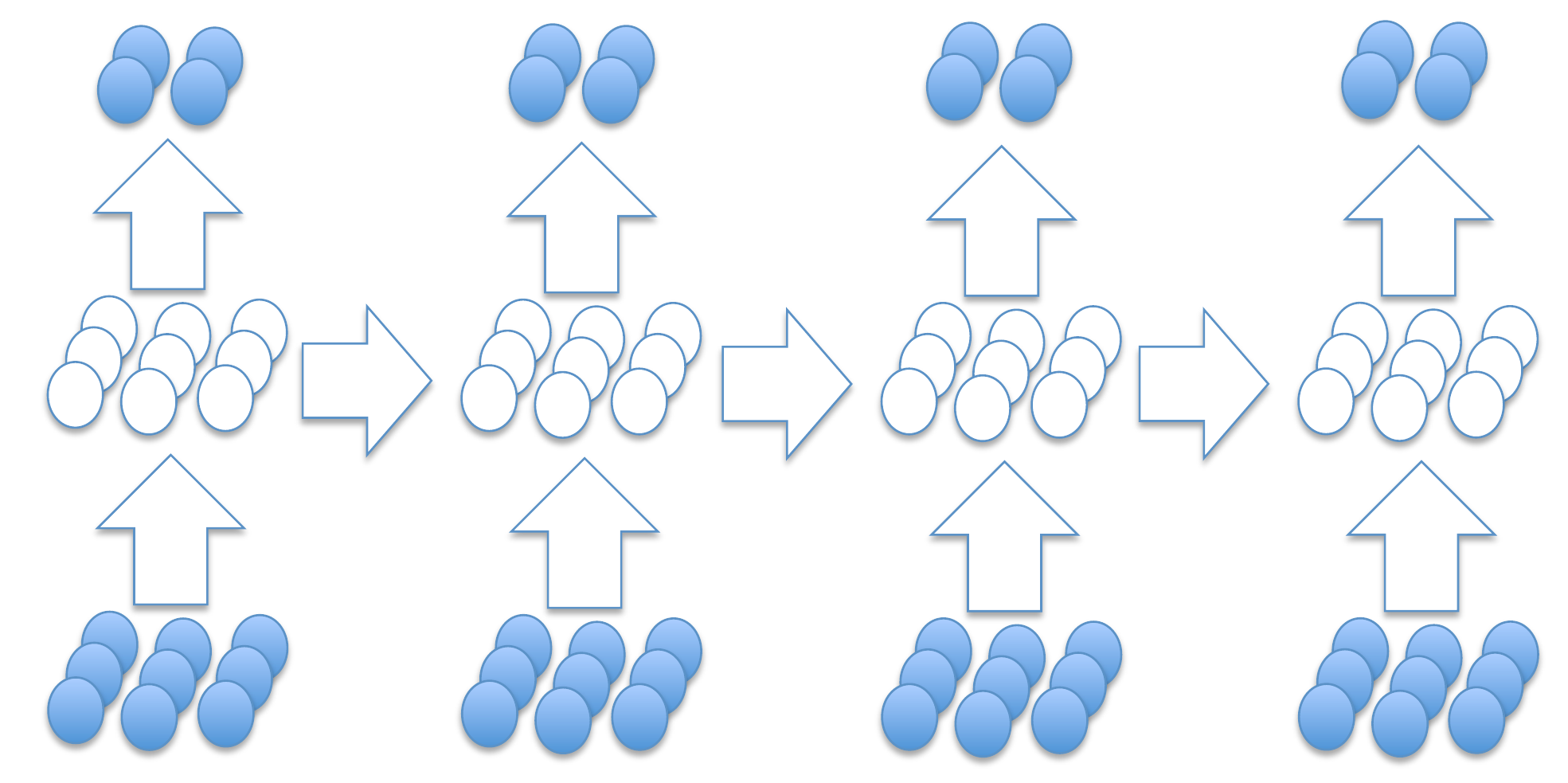}
\par\end{centering}
\caption{Matrix recurrent neural nets. Circles represent neurons, arrows represent
the directions of influence. \label{fig:Matrix-recurrent-neural}}
\end{figure}

In this paper, we formalize those separate ideas by deriving \emph{a
new common building block of neuron networks in that} \emph{all the
neuron representations are matrices}. We replace Eq.~(\ref{eq:vector-neuron})
by the following:
\begin{equation}
Y=\sigma\left(U^{\intercal}XV+B\right)\label{eq:matrix-neuron}
\end{equation}
where $X,Y$ are matrix representation; $U,V,B$ are parameters. While
this idea has been suggested in a contemporary work of \cite{gao2017matrix},
our work aims to be more thorough and systematic. \uline{}First,
we generalize state-of-the-art feed-forward neural networks and recurrent
neural networks (RNNs) to accommodate matrix forms (see Fig.~\ref{fig:Matrix-recurrent-neural}
for an illustration of a matrix RNN). Second, we demonstrate how to
formulate recent memory-augmented networks using the matrix representation
with appropriate choice of weights $U$ and $V$ in Eq.~(\ref{eq:matrix-neuron}).
Third, we show that several recent neural graph models are actually
instances of matrix nets, and how matrix representation leads to expressive
extensions to include multiple attentions. 

Moreover, to prove the advantages of this new representation, we
have designed a comprehensive suite of five different experiments.
In the first two, we explore the learning curves of matrix feed-forward
nets and the reconstruction capability of matrix auto-encoders to
support an argument that the matrix mapping in Eq.~(\ref{eq:matrix-neuron})
does provide a so-called \textit{``structural regularization''}
which cannot be seen on the vector mapping. In the next two, we test
the performance of matrix recurrent nets on 2 tasks: sequence-to-sequence
learning and classification of EEG signals. Our purpose is to assert
that matrix recurrent nets with more memory but far fewer parameters
can easily outperform the vector counterparts. In the last experiment,
we demonstrate matrix-based graph models on node classification for
citation networks, showing how matrix representation improves upon
vector representation, and how matrix gives rise to multi-attention
mechanisms that offers further improvement.

\section{Matrix Networks \label{sec:Methods}}

We design matrix-based neural networks where the input, output, hidden
state and memory are matrices. Let us start by introducing some notations:

\begin{eqnarray}
\text{mat}_{1}(P;\thetab_{1}) & \coloneqq & U^{\intercal}PV+B\label{eq:mat-mapping-1}
\end{eqnarray}
where $P\in\Real^{r\times c}$ is a neuron matrix of $r$ rows and
$c$ columns, $U\in\Real^{r\times r'}$ and $V\in\Real^{c\times c'}$
are called \textit{row-mapping} and \textit{column-mapping} matrices
respectively, $B\in\Real^{r'\times c'}$ is a bias matrix and$\thetab_{1}=\{U,V,B\}$
denotes parameters of the model. 

Next, we will show how to derive (i) multi-attention mechanisms (Sec.~\ref{subsec:Memory-Augmented-Nets})
(ii) new matrix-based deep feed-forward nets (Sec.~\ref{subsec:Matrix-Feedforward-Nets}),
and (iii) new matrix recurrent nets (Sec.~\ref{subsec:Matrix-RNN}).

\subsection{Multi-Attention and Memory \label{subsec:Memory-Augmented-Nets}}

In Eq.~(\ref{eq:mat-mapping-1}), given a matrix $P$ of $r$ rows,
we can soft-attend to its rows by requiring $U=\left[\alphab_{1},\alphab_{2}...\alphab_{k}\right]$,
where each $\alphab_{j}$ is an attention vector subject to $\sum_{i=1}^{m}\alpha_{ij}=1$
and $\alpha_{ij}\ge0$ for $j=1,2,..,k$. In this setting, each read
pass returns an aggregated vector:
\begin{equation}
\sb_{j}=\alphab_{j}^{\intercal}P\label{eq:attention}
\end{equation}
which is then transformed using the transformation matrix $V$. Typically,
$\alphab_{j}$ is parameterized using a neural network with softmax
activation, possibly as a function of $P$ and its context (if available). 

Indeed, several memory\textendash augmented networks such as Neural
Turing Machine \cite{graves2014neural}, Memory Network \cite{weston2014memory},End-to-End
Memory Network \cite{sukhbaatar2015end} can be formulated in similar
ways with $P$ being the external memory and $U$ is the collection
of read heads. 

\subsection{Matrix Feed-forward Nets \label{subsec:Matrix-Feedforward-Nets}}

The one-layer matrix feed-forward net that maps a matrix input $X$
onto a matrix output $Y$ (\emph{mat2mat}) is readily defined as:
\begin{align}
Y & =\sigma_{Y}\left(\mat_{1}\left(X;\thetab_{Y}\right)\right)\label{eq:feedforward-net}
\end{align}
Deep models can be generalized as usual by stacking multiple hidden
matrix layers. To improve the fast credit assignment for very deep
nets, we can employ \emph{skip-connections}, just like the case of
vector-based nets. In vector representation, it typically takes the
following form:
\[
\hb_{t}=\rb_{t}\odot\hb_{t-1}+\zb_{t}\odot\tilde{\hb_{t}}
\]
where $\hb_{t}$and $\tilde{\hb_{t}}$ are the final and intermediate
representation at step $t$ (assuming $\hb_{0}$ is the input vector
$\xb$), $\rb_{t}\in\left[\boldsymbol{0},\boldsymbol{1}\right]$,
$\zb_{t}\in\left[\boldsymbol{0},\boldsymbol{1}\right]$ are gates,
and $\odot$ is point-wise multiplication. The layer $t-1$ is said
to be skip-connected\footnote{The skip-connections can indeed go further down many times and many
steps to $t-k$, for $k=1,2...,t-1$, e.g., see \cite{huang2017densely}.} to layer $t$ through the term $\rb_{t}\odot\hb_{t-1}$. The extension
to matrix is straightforward:
\[
H_{t}=R_{t}\odot H_{t-1}+Z_{t}\odot\tilde{H_{t}}
\]
For example, the Highway Network \cite{srivastava2015training} can
be extended as:
\begin{eqnarray*}
Z_{t} & = & \text{sigm}\left(\text{mat}_{1}\left(H_{t-1};\thetab_{t,z}\right)\right)\\
\tilde{H}_{t} & = & \sigma\left(\text{mat}_{1}\left(H_{t-1};\thetab_{t,h}\right)\right)\\
H_{t} & = & (1-Z_{t})\odot H_{t-1}+Z_{t}\odot H_{t-1}
\end{eqnarray*}
Here $R_{t}=1-Z_{t}$. The matrix ResNet \cite{he2016deep} is similar:
$H_{t}=H_{t-1}+F\left(H_{t}\right)$ where $F\left(H_{t}\right)$
is a residual feedforward subnet that map a matrix $H_{t}$ into another
matrix $\tilde{H_{t}}$ in the same space. These networks can be made
recurrent by typing parameters across layers.

\subsection{Matrix Recurrent Nets \label{subsec:Matrix-RNN}}

The standard vector-based recurrent neural network is a mapping from
a vector sequence to another sequence. The matrix alternative thus
maps an input matrix sequence $X_{1:T}$ to an output matrix sequence
$Y_{1:T}$, via a hidden matrix sequence $H_{1:T}$, as follows. Let:

\begin{eqnarray}
\mat_{2}(P,Q;\thetab_{2}) & \coloneqq & U_{p}^{\intercal}PV_{p}+U_{q}^{\intercal}QV_{q}+B\label{eq:mat-mapping-2}
\end{eqnarray}
where $P,Q$ are neuron matrices and $\thetab_{2}=\{U_{p},V_{p},U_{q},V_{q},B\}$
are parameters. The network dynamics are summarized in the following
equations:

\begin{align}
Y_{t} & =\sigma_{Y}\left(\mat_{1}\left(H_{t};\thetab_{Y}\right)\right)\label{eq:Mat-RNN-out}\\
H_{t} & =\sigma_{H}\left(\mat_{2}\left(X_{t},H_{t-1};\thetab_{H}\right)\right)\label{eq:Mat-RNN-state}
\end{align}
The number of parameters is roughly quadratic in the dimensions of
matrices $X_{t},H_{t},Y_{t}$. This is inline with the capacity of
the short-term memory stored in $H_{t}$ . Thus the parameter\textendash memory
ratio is a constant, unlike the ratio in the classical case of vector
memory, which grows linearly with vector size.

The generalization from vanilla matrix RNN to matrix LSTM is straightforward.
The formula of a LSTM block (without peep-hole connection) at time
step $t$ can be specified as follows:

\begin{eqnarray*}
I_{t} & = & \text{sigm}(\mat_{2}(X_{t},H_{t-1};\thetab_{i}))\\
F_{t} & = & \text{sigm}(\mat_{2}(X_{t},H_{t-1};\thetab_{f}))\\
O_{t} & = & \text{sigm}(\mat_{2}(X_{t},H_{t-1};\thetab_{o}))\\
\hat{C_{t}} & = & \text{tanh}(\mat_{2}(X_{t},H_{t-1};\thetab_{c}))\\
C_{t} & = & F_{t}\odot C_{t-1}+I_{t}\odot\hat{C}_{t}\\
H_{t} & = & O_{t}\odot\text{tanh}(C_{t})
\end{eqnarray*}
where $\odot$ denotes the Hadamard product; and $I_{t}$, $F_{t}$
$O_{t}$, $\hat{C_{t}}$ are input gate, forget gate, output gate
and cell gate at time $t$, respectively. Note that the memory cells
$C_{t}$ that store, forget and update information is a matrix.

To save space for other discussion, we will not present the formula
of the matrix GRU here since it can easily be derived from the formulas
of vector GRU in the same way as LSTM.

\section{Matrix Representation of Graphs \label{sec:Matrix-Rep-of-Graph}}

In this section, we show \textit{how natural the matrix formulation
is for deep graph modeling}. In particular, we focus our attention
on the work in \cite{scarselli2009graph,li2016gated,pham2017column}.
We will use the notation from \cite{pham2017column}, where the network
is called Columm Network (CLN). For the generalization to other types
of graph neural networks \cite{defferrard2016convolutional,kipf2016semi}
based on spectral graph theory, please refer to Appx.~\ref{subsec:Graph-Convolution-as}. 

Let $\Graph=(\Vertices,\Edges)$ be a graph, where $\Vertices$ is
a set of nodes and $\Edges$ is a set of edges. In CLN, each node
in $\Vertices$ is modeled using a deep feed-forward net such as Highway
Network \cite{srivastava2015training}. Different from standard feed-forward
nets, here all nets are inter-connected where information is passed
between nets along the edges defined by $\Edges$. 

Denote by $\hb_{i}^{t}\in\Real^{1\times d}$ the activation vector
of node $i$ at step $t$. CLN computes $\tilde{\hb}_{i}^{t}$, an
aggregate of neighbor states at step $t$, where the neighborhood
is defined by $\Edges$:
\begin{equation}
\tilde{\hb}_{i}^{t}=\frac{1}{\left|\mathcal{N}(i)\right|}\sum_{j\in\mathcal{N}(i)}\hb_{j}^{t}\label{eq:vec-CLN-aggregate}
\end{equation}
where $\mathcal{N}(i)$ is the neighborhood of node $i$, as defined
by $\Edges$. Then the activation is a function of the previous step
as follows:
\begin{equation}
\hb_{i}^{t}=f\left(\hb_{i}^{t-1},\tilde{\hb}_{i}^{t-1}\right)\label{eq:vec-CLN-update}
\end{equation}

We now show how CLN can be represented using matrix notation. First,
all hidden states in step $t$ can be stacked vertically (by row)
to form a matrix $H^{t}=\left[\hb_{1}^{t}\mid\hb_{2}^{t}\mid...\mid\hb_{|\Vertices|}^{t}\right]$,
i.e., $H^{t}\in\Real^{|\Vertices|\times d}$. Likewise, let $\tilde{H}^{t}=\left[\tilde{\hb}_{1}^{t}\mid\tilde{\hb}_{2}^{t}\mid...\mid\tilde{\hb}_{|\Vertices|}^{t}\right]$.

Now let $A$ be the adjacency matrix of graph $G$ and $\tilde{A}$
be the normalized version of $A$, that is $\tilde{A}_{ij}=\frac{1}{\left|\mathcal{N}(i)\right|}$
for $j\in\mathcal{N}(i)$ and $\tilde{A}_{ij}=0$ for $j\notin\mathcal{N}(i)$.
Eq.~(\ref{eq:vec-CLN-aggregate}) can be written as:
\begin{equation}
\tilde{H}^{t}=\tilde{A}H^{t}\label{eq:matrix-CLN-aggregate}
\end{equation}
and Eq.~(\ref{eq:vec-CLN-update}) can be rewritten as:
\begin{equation}
H^{t}=F\left(H^{t-1},\tilde{H}^{t-1}\right)\label{eq:matrix-CLN-update}
\end{equation}
where $F$ is an arbitrary matrix neural network.

\subsection{Column Networks with Multi-Attentions \label{subsec:Multiple-Attention-Column-Networ}}

Putting Eq.~(\ref{eq:matrix-CLN-aggregate}) in the context of Sec.~\ref{subsec:Memory-Augmented-Nets}
immediately suggests that $\Lambda$ can be extended to stimulate
\emph{attention mechanism} rather than simple averaging. The normalized
adjacency matrix $\tilde{A}$ now becomes the attention matrix $\Lambda$
whose element $\Lambda_{ij}>0$ is the probability that that node
$i$ chooses to include information from node $j$. 

In our experiment in Sec.~\ref{subsec:Exp-Multiple-Attention}, we
use the following attention formula: 

\begin{eqnarray*}
\Lambda_{ij} & = & \softmax\ g(\hb_{i}^{t-1},\hb_{j}^{t-1})
\end{eqnarray*}
with $g$ is a bilinear neural network: $g(\xb,\yb)=\xb^{\intercal}W\yb+\ab^{\intercal}\xb+\bb^{\intercal}\yb+\cb$.
The appear of parameter $W$ in $g$ allow the interaction between
a node $i$ and its neighbor $j$ to be captured.

Sometimes, it is not enough to aggregate all neighboring states of
a node $i$ into a single vector, especially when the size of $\Neigh(i)$
is big (e.g. in a citation network, one paper can be cited by hundreds
or thousands of other papers). Therefore, we propose an architecture
called \textit{``multi-attention''} where we compute $n$ attention
matrices $\Lambda^{(1)},...,\Lambda^{(n)}$ using $n$ different neural
networks $g^{(1)},...,g^{(n)}$. Replace $\tilde{A}$ in Eq.~(\ref{eq:matrix-CLN-aggregate})
by $\Lambda^{(i)}$ ($i=\overline{1,n}$), we have: 

\begin{eqnarray}
\tilde{H}^{(i)t} & = & \Lambda^{(i)}H^{t}\ \ \ \forall i=\overline{1,n}\label{eq:matrix-CLN-aggregate-1}
\end{eqnarray}
and the final $\tilde{H}^{t}\in\Real^{|\Vertices|\times nd}$ is the
concatenation of $\tilde{H}^{(i)t}$ ($i=\overline{1,n}$) by columns:

\[
\tilde{H}^{t}=\left[\tilde{H}^{(1)t},\tilde{H}^{(2)t},...,\tilde{H}^{(n)t}\right]
\]
Since $\tilde{H}^{t}$ still has matrix form (only its shape is modified),
the generic formula in Eq.~(\ref{eq:matrix-CLN-update}) remains
unchanged.

\section{Experimental Results}

In this section, we present experimental results on validating the
matrix neural architectures described in Section.~\ref{sec:Methods}.
Our primary purpose is to demonstrate that when matrix-like structures
are present, matrix nets will clearly outperform the vector nets.
For all experiments presented below, our networks use ReLU activation
function and are trained using Adam \cite{kingma2014adam} (with default
settings). 

\subsection{Learning Characteristics of Matrix Neural Networks}

We first evaluate the learning curves for deep matrix feed-forward
nets (FFNs) under various settings. For convenience, we use handwritten
digit and facial images and treat an image as a matrix, ignoring its
translation- and rotation-invariant nature. With these in mind, images
can be either row- or column-permuted before feeding to matrix nets,
and these rule out applying standard CNN for feature extraction.

\subsubsection{MNIST}

The MNIST dataset consists of 70K handwritten digits of size $28\times28$,
with 50K images for training, 10K for testing and 10K for validation.
To test the ability to accommodate very deep nets without skip-connections
\cite{srivastava2015training}, we create vector and matrix FFNs with
increasing depths. The top layers are softmax as usual for both vector
and matrix nets. We compare matrix nets with the hidden shape of $20\times20$
and $50\times50$ against vector nets containing 50, 100 and 200 hidden
units.

\begin{figure}[h]
\begin{centering}
\subfigure[Vector nets]{
	\label{fig:MNIST-vec30}
	\includegraphics[height=0.4\textwidth]{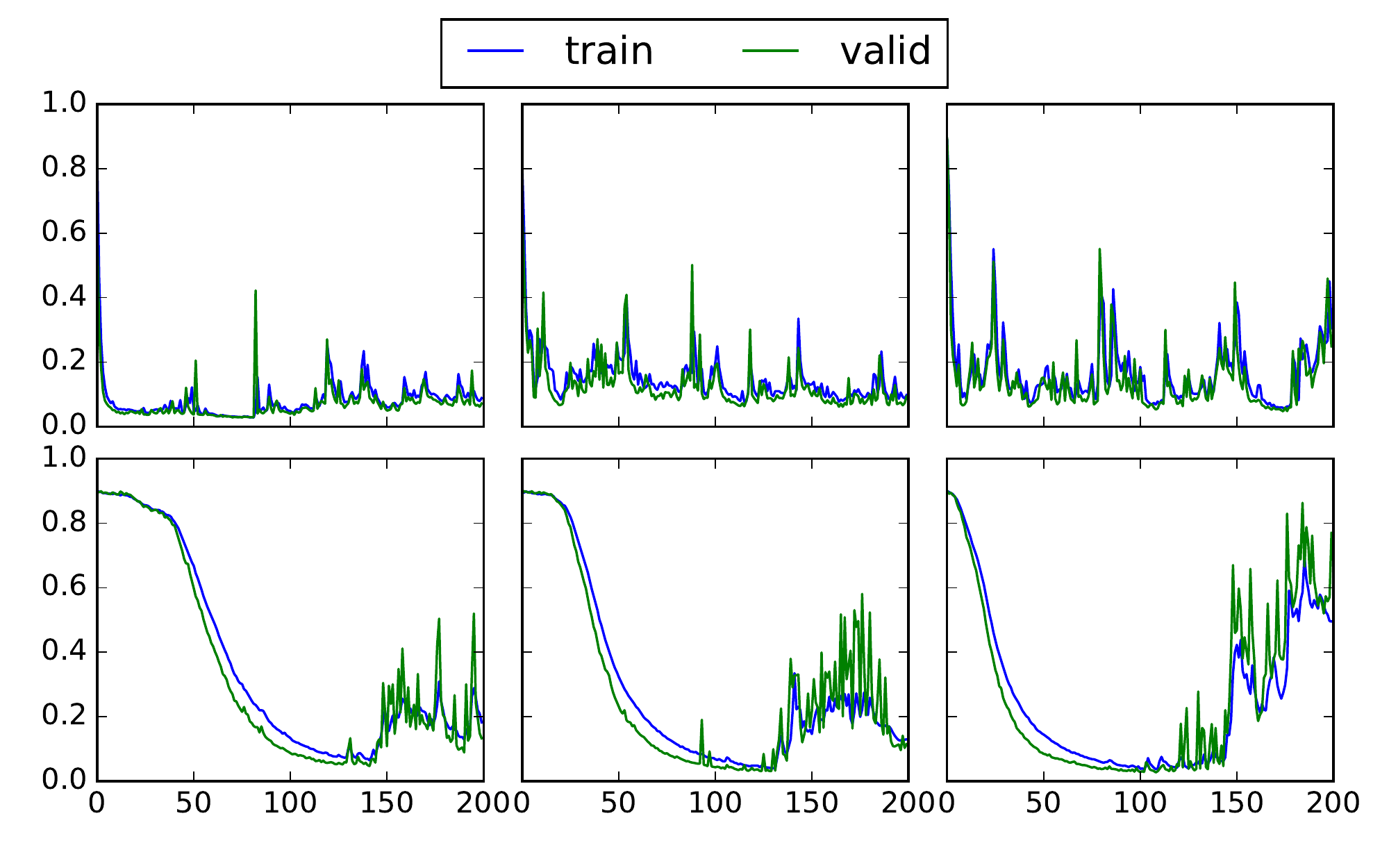}
}
\subfigure[Matrix nets]{
    \label{fig:MNIST-mat30}
	\includegraphics[height=0.4\textwidth]{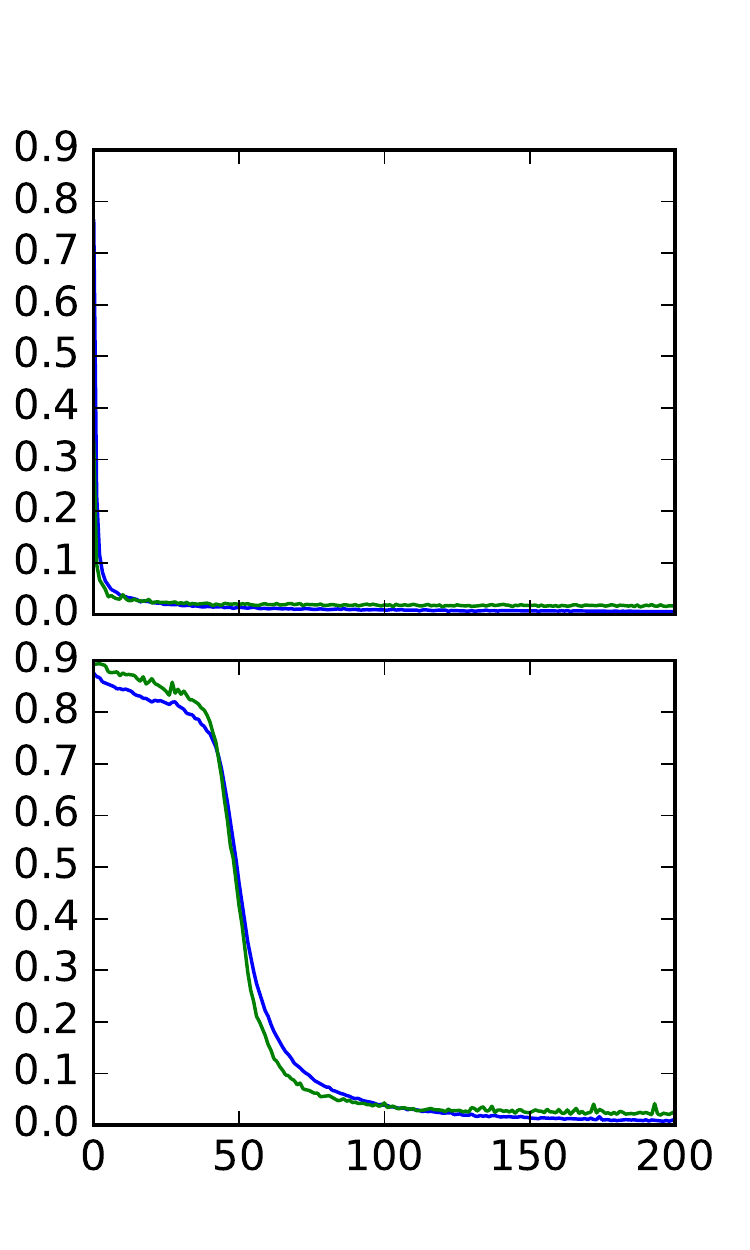}
}
\par \end{centering}
\caption{Learning curves of vector and matrix feed-forward nets over MNIST.\textbf{
(a)} \textbf{Left to right:} Vector nets with $50$, $100$, $200$
hidden units. \textbf{(b)}: Matrix nets with $50\times50$ hidden
units. \textbf{Top:} $30$-layer nets without Batch Norm. \textbf{Bottom:}
$50$-layer nets with Batch Norm.\label{fig:MNIST_err_curve}}
\end{figure}

We observe that without Batch-Norm (BN) \cite{ioffe2015batch}, vector
nets struggle to learn when the depth goes beyond $20$, as expected.
The erratic learning curves of the vector nets at depth 30 are shown
in Fig.~\ref{fig:MNIST_err_curve}(a), top row. With the help of BN layers,
the vector nets can learn normally at depth 30, but again fail beyond
depth 50 (see Fig.~\ref{fig:MNIST_err_curve}(a), bottom row). The matrix
nets are far better: They learn smoothly at depth 30 without BN layers
(Fig.~\ref{fig:MNIST_err_curve}(b), top). With BN layers, they still learn
well at depth 50 (Fig.~\ref{fig:MNIST_err_curve}(b), bottom) and can manage
to learn up to depth 70 (result is not shown here).

\begin{figure}[h]
\begin{centering}
\includegraphics[width=0.7\columnwidth]{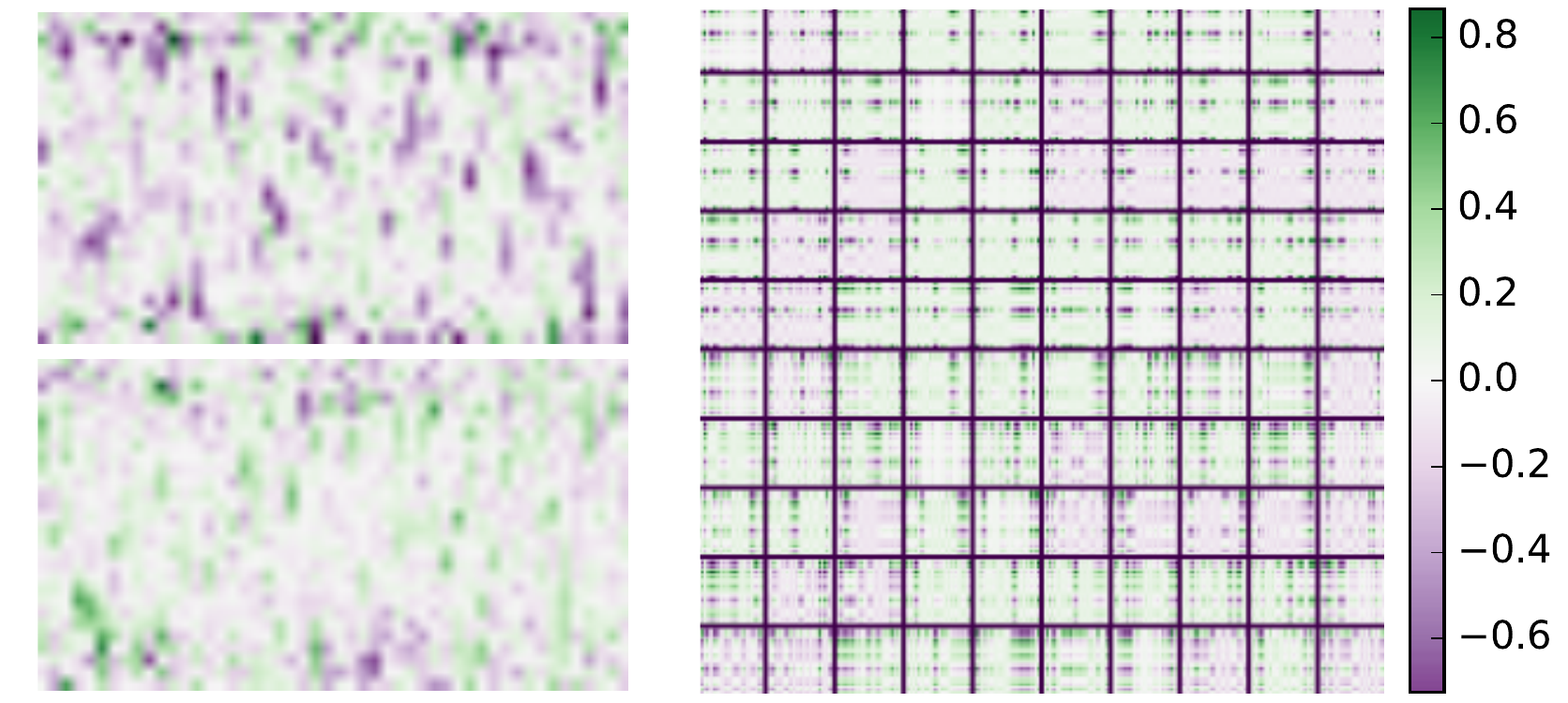}
\par \end{centering}
\caption{The normalized weights of the first layer of the 30-layer matrix feed-forward
net with hidden size of $50\times50$ trained on MNIST. Weights of
other layers look quite similar. \textbf{Top left:} Row mapping matrix
$U$ of shape $28\times50$. \textbf{Bottom left:} Column mapping
matrix $V$ of shape $28\times50$. \textbf{Right:} Each square tile
$i$, $j$ show the weight $W_{:,i\times50+j}$ which is the outer
product of $U_{:,i}$ and $V_{:,j}$.\label{fig:MNIST_weights}}
\end{figure}

We visualize the weights of the first layer of the matrix net with
hidden layers of $50\times50$ (the weights for $20\times20$ layers
are similar) in Fig.~\ref{fig:MNIST_weights} for a better understanding.
In the plots of $U$ and $V$ (top and bottom left of Fig.~\ref{fig:MNIST_weights},
respectively), the short vertical brushes indicate that some adjacent
input features are highly correlated along the row or column axes.
For example, the digit $1$ has white pixels along its line which
will be captured by $U$. In case of $W$, each square tile in Fig.~\ref{fig:MNIST_weights}(right)
corresponds to the weights that map the entire input matrix to a particular
element of the output matrix. These weights have cross-line patterns,
which differ from stroke-like patterns commonly seen in vector nets. 

\subsubsection{Matrix Autoencoders for Corrupted Face Reconstruction}

To evaluate the ability of learning structural information in images
of matrix neural nets, we conduct experiments on the Frey Face dataset\footnote{http://www.cs.nyu.edu/\textasciitilde{}roweis/data.html},
which consists of 1,965 face images of size $28\times20$, taken from
sequential frames of a video. We randomly select 70\% data for training,
20\% for testing and 10\% for validation. Test images are corrupted
with $5\times5$ black square patches at random positions. Auto-encoders
(AEs) are used for this reconstruction task. We build deep AEs consisting
of 20 and 40 layers. For each depth, we select vector nets with 50,
100, 200 and 500 hidden units and matrix nets with hidden shape of
$20\times20$, $50\times50$, $100\times100$ and $150\times150$.
The encoders and the decoders have tied weights. The AEs are trained
with backprop, random noise added to the input with ratio of $0.2$,
and L1 and L2 regularizers. 

\begin{figure*}
\begin{centering}
\subfigure[]{
	\label{fig:FreyFace_a}
	\begin{centering}
		\includegraphics[height=0.33\textwidth]{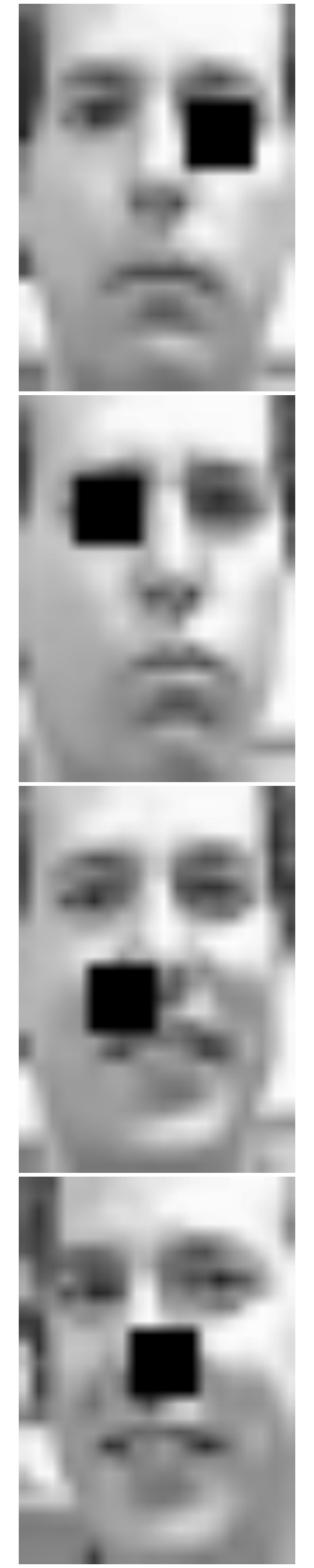}
		\includegraphics[height=0.33\textwidth]{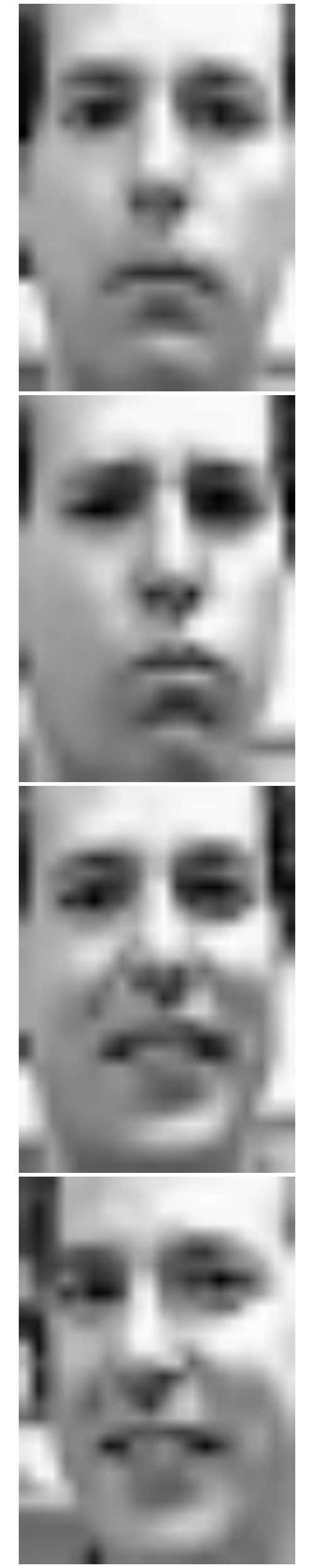}
	\end{centering}
}
\subfigure[]{
	\label{fig:FreyFace_b}
	\begin{centering}
		\includegraphics[height=0.33\textwidth]{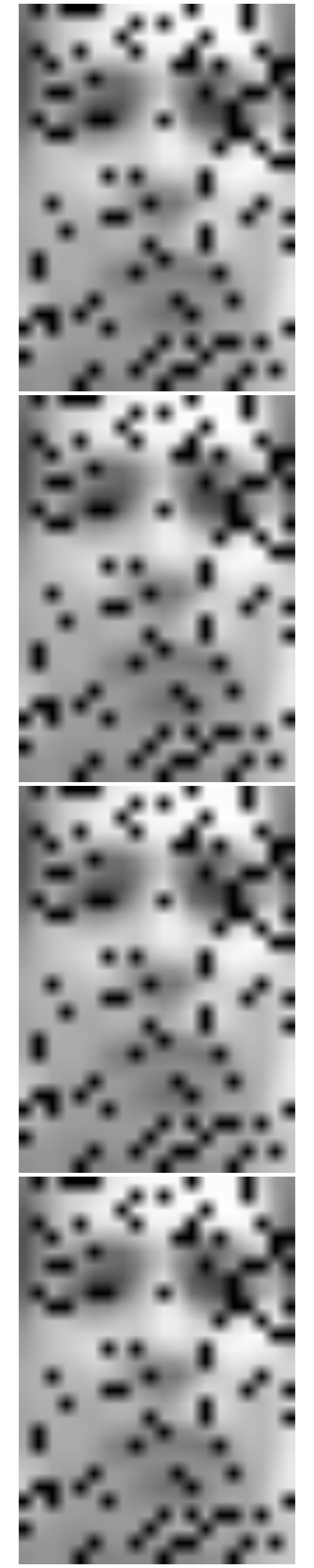}
		\includegraphics[height=0.33\textwidth]{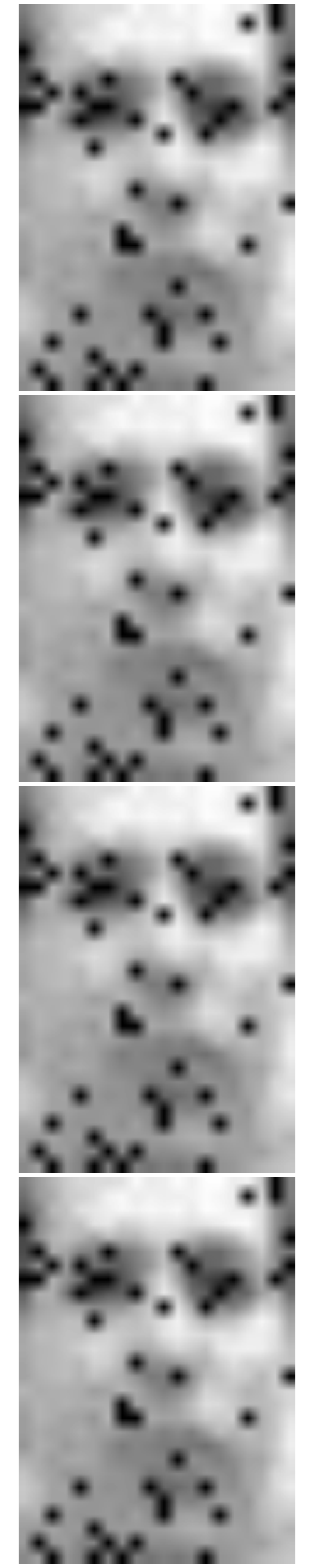}
	\end{centering}
}
\subfigure[]{
	\label{fig:FreyFace_c}
	\begin{centering}
		\includegraphics[height=0.33\textwidth]{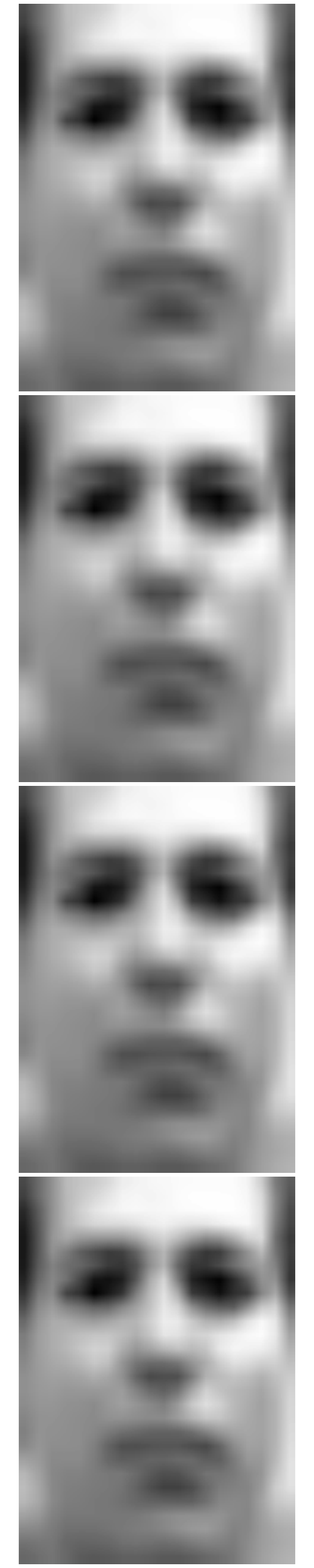}
	\end{centering}
}
\subfigure[]{
	\label{fig:FreyFace_d}
	\begin{centering}
		\includegraphics[height=0.33\textwidth]{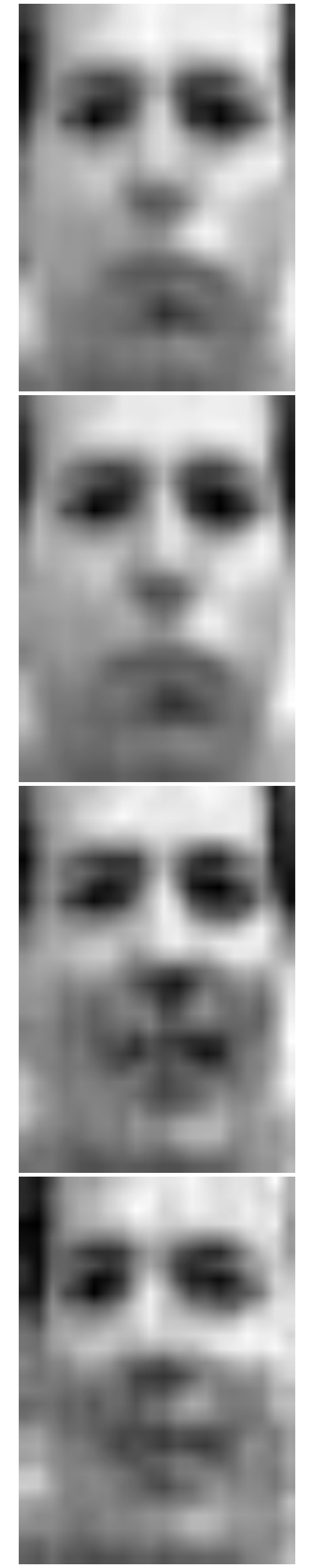}
		\includegraphics[height=0.33\textwidth]{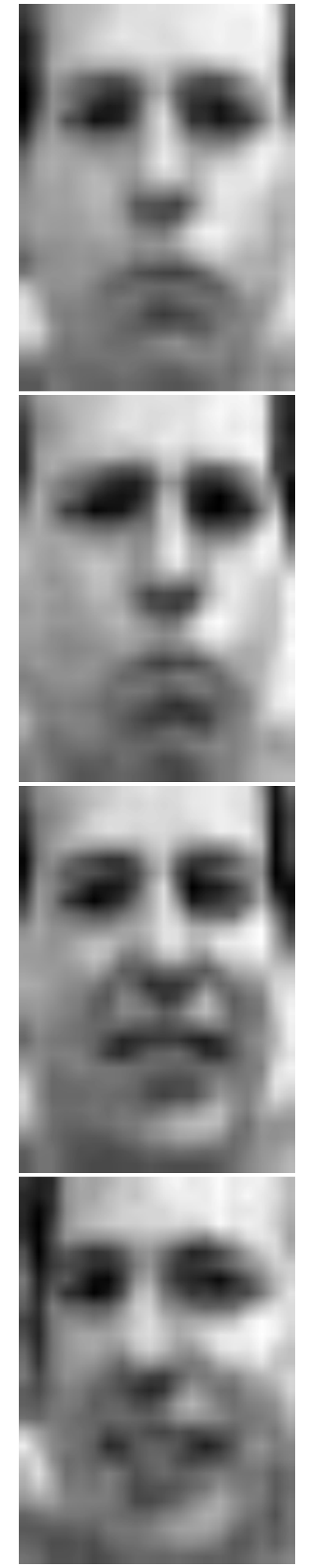}
		\includegraphics[height=0.33\textwidth]{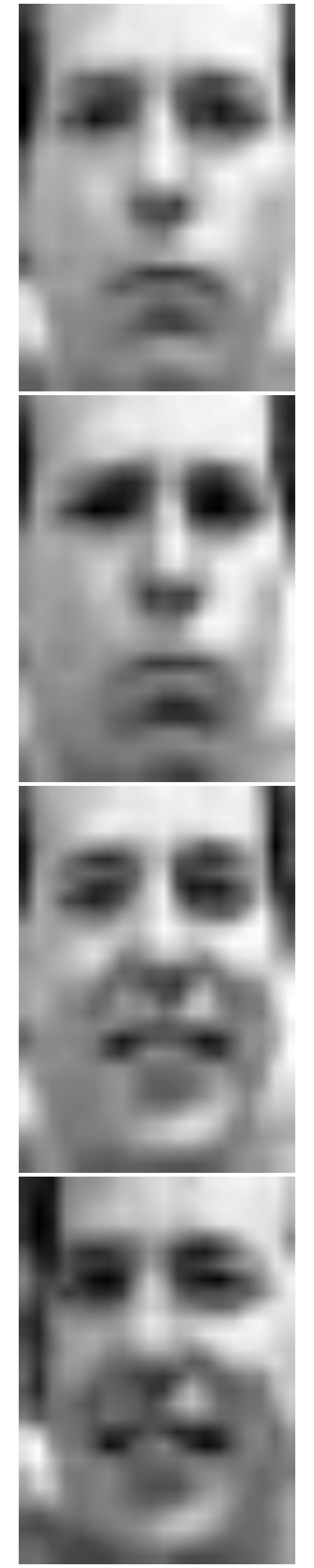}
	\end{centering}
}
\subfigure[]{
	\label{fig:FreyFace_e}
	\begin{centering}
		\includegraphics[height=0.33\textwidth]{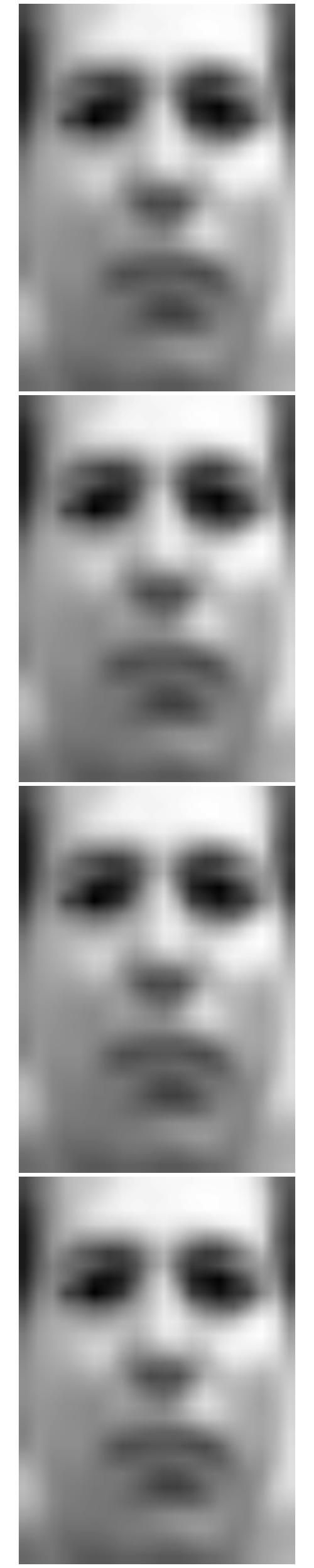}
	\end{centering}
}
\subfigure[]{
	\label{fig:FreyFace_f}
	\begin{centering}
		\includegraphics[height=0.33\textwidth]{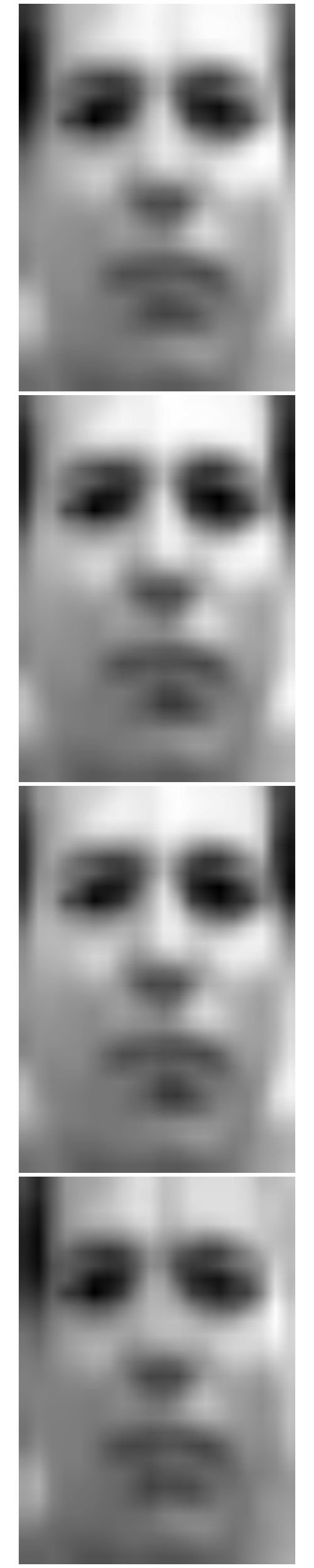}
		\includegraphics[height=0.33\textwidth]{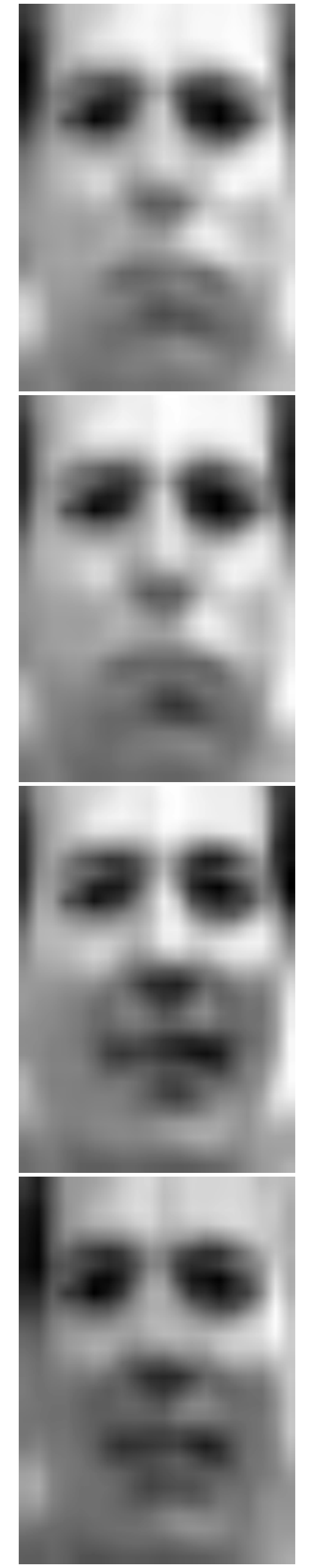}
	\end{centering}
}
\par \end{centering}
\caption{Images reconstructed by matrix AEs and vector AEs in comparison with
ground truth. All AEs from \textbf{(b)} to \textbf{(e)} have 40 layers
while the AEs at \textbf{(f)} have 20 layers.\textbf{ (a):} corrupted
inputs and ground truth; \textbf{(b):} vector AEs (hidden size of
$50$ and $500$) without regularization; \textbf{(c):} vector AEs
with regularization; \textbf{(d):} matrix AEs (hidden size of $20\times20$,
$50\times50$ and $150\times150$) without regularization; \textbf{(e):}
matrix AEs with L1 regularization; \textbf{(f):} matrix AEs (hidden
size of $50\times50$ and $150\times150$) with L2 regularization.\label{fig:FreyFace_rec} }
\end{figure*}

Once trained, AEs are used to reconstruct the test images. Fig.~\ref{fig:FreyFace_rec}
presents several reconstruction results. Vector AEs fail to learn
to reconstruct either with or without weight regularization. Without
weight regularization, vector AEs fail to remove noise from the training
images (Fig.~\ref{fig:FreyFace_b}), while with weight regularization
they collapse to a single mode (Fig.~\ref{fig:FreyFace_b})\footnote{This happens for all hidden sizes and all depth values of vector AEs
specified above.}. Matrix AEs, in contrast, can reconstruct the test images quite well
\emph{without weight regularization} (see Fig.~\ref{fig:FreyFace_d}).
In fact, adding weight regularization to matrix AEs actually deteriorates
the performance, as shown in Figs.~\ref{fig:FreyFace_rec}(e,f).
This suggests an expected behavior in which matrix-like structures
in images are preserved in matrix neural nets, enabling missing information
to be recovered.

\subsection{Sequence to Sequence Learning with Moving MNIST}

In this experiment, we compare the performance of matrix and vector
recurrent nets in a \emph{sequence-to-sequence} (seq2seq) learning
task \cite{sutskever2014sequence,srivastava2015unsupervised}. We
choose the Moving MNIST dataset\footnote{http://www.cs.toronto.edu/\textasciitilde{}nitish/unsupervised\_video/}
which contains 10K image sequences. Each sequence has length of 20
showing 2 digits moving in $64\times64$ frames. We randomly divide
the dataset into 6K, 3K and 1K image sequences with respect to training,
testing and validation. In our seq2seq model, the encoder and the
decoder are both recurrent nets. The encoder captures information
of the first 15 frames while the decoder predicts the last 5 frames
using the hidden context learnt by the encoder. Different from \cite{srivastava2015unsupervised},
the decoder do not have readout connections\footnote{the predicted output of the decoder at one time step will be used
as input at the next time step} for simplicity. We build vector seq2seq models with hidden sizes
ranging from $100$ to $2000$ for both the encoder and the decoder.
In case of matrix seq2seq models, we choose hidden sizes from from
$10\times10$ to $200\times200$. Later in this section, we write
vector RNN/LSTM to refer to a vector seq2seq model with the encoder
and decoder are RNNs/LSTMs. The same notation applies to matrix.

\begin{figure}[t]
\begin{centering}
\includegraphics[width=0.6\textwidth]{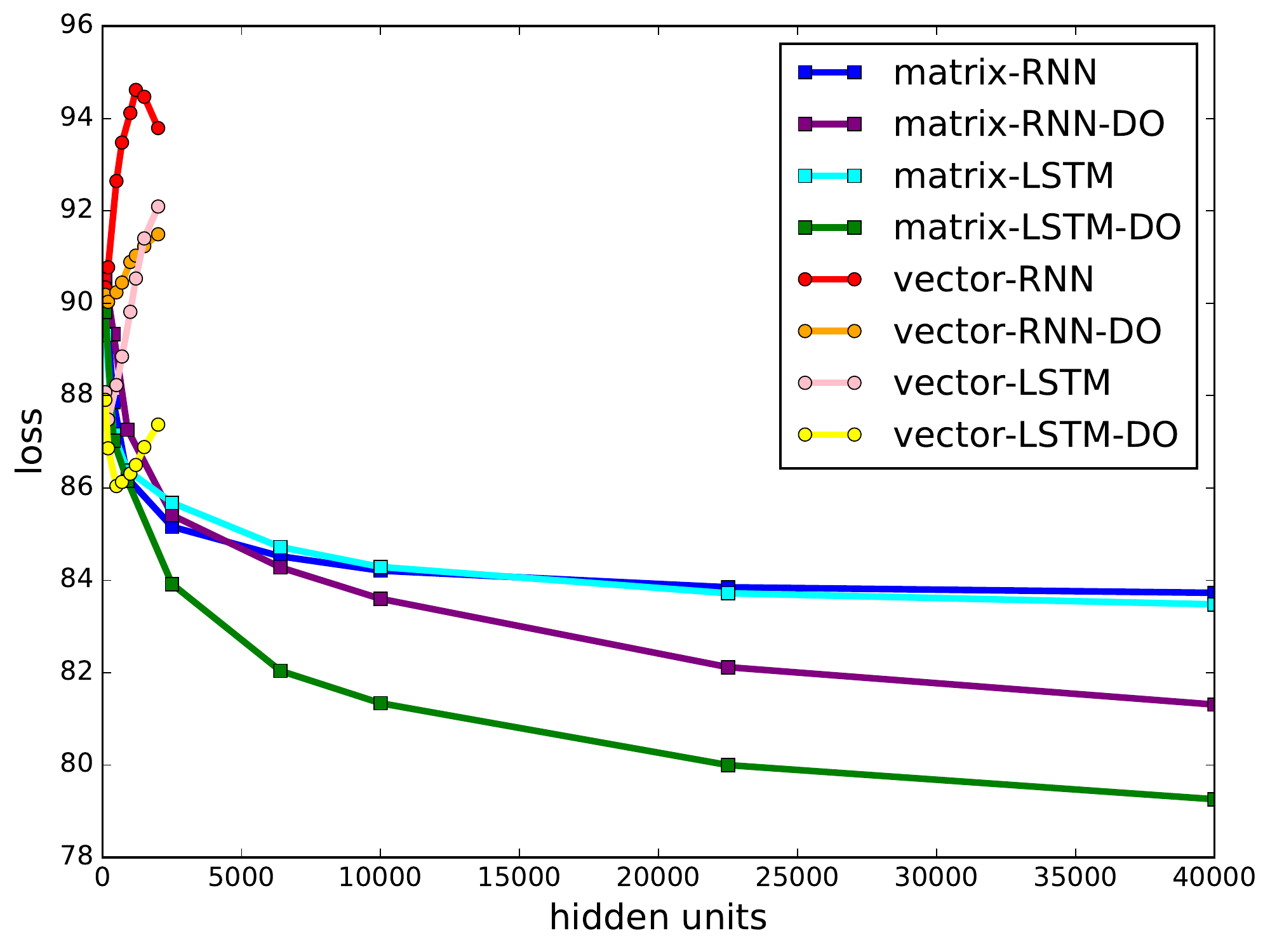}
\par \end{centering}
\caption{Reconstruction loss over test data of matrix and vector seq2seq models
as functions of hidden unit. \emph{DO} is short for Dropout which
is set at 0.2 for inputs and 0.5 for hidden units. Best viewed in
color.\label{fig:MovingMNIST_plot}}
\end{figure}

It is important to emphasize that matrix nets are far more compact
than the vector counterparts. For example, the vector RNNs require
nearly 30M parameters for 2K hidden units while the matrix RNNs only
need about 400K parameters (roughly 75 times fewer) but have 40K hidden
units (20 times larger)\footnote{For LSTMs, the number of parameters quadruples but the relative compactness
between vector and matrix nets remain the same.}. The parameter inflation exhibits a huge redundancy in vector representation
which makes the vector nets susceptible to overfitting. Therefore,
after a certain threshold (200 in this case), increasing the hidden
size of a vector RNN/LSTM will deteriorate its performance. Matrix
nets, in contrast, are consistently better when the hidden shape becomes
larger, suggesting that overfitting is not a problem. Remarkably,
a matrix RNN/LSTM with hidden shape of $50\times50$ is enough to
outperform vector RNNs/LSTMs of any size with or without dropout (see
Fig.~\ref{fig:MovingMNIST_plot}). Dropout does improve the representations
of both vector and matrix nets but it cannot eliminate the overfitting
on the big vector nets.

\subsection{Sequence Classification with EEG}

\begin{figure}
\begin{centering}
\includegraphics[width=0.8\textwidth]{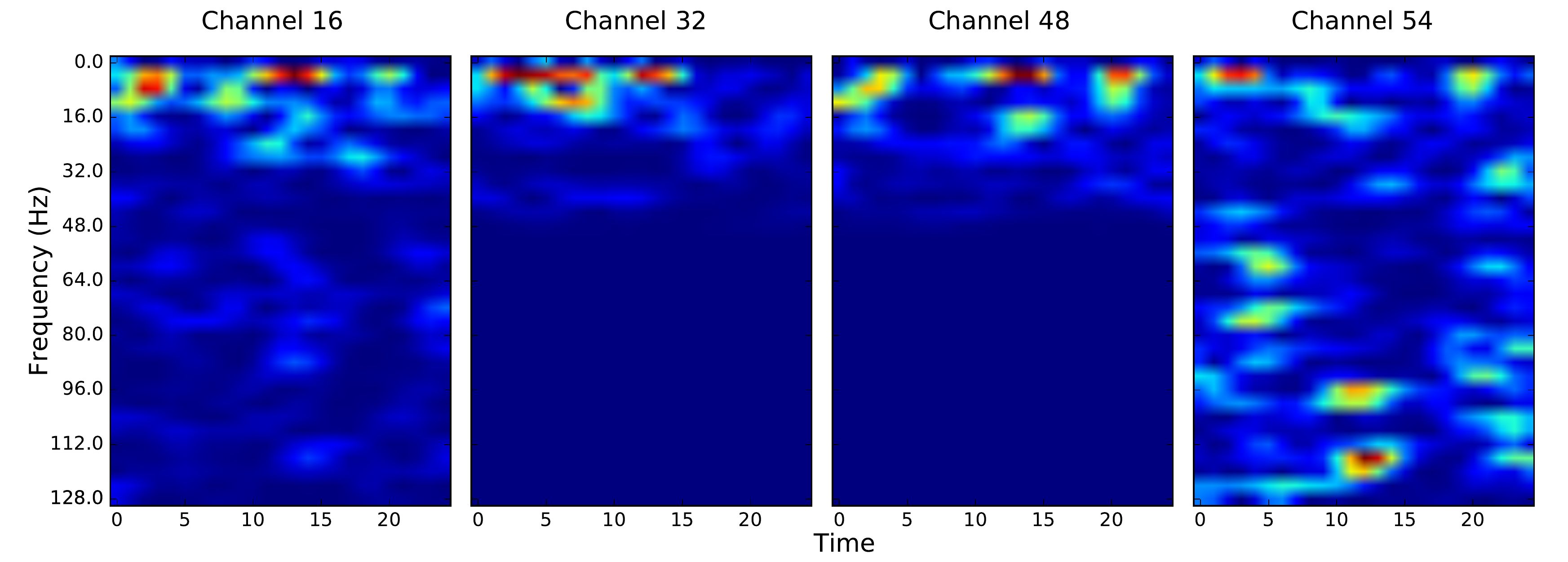}
\par \end{centering}
\caption{Spectrograms of four channels taken from one trial of a random alcoholic
subject. Best viewed in color.\label{fig:EEG_Spec}}
\end{figure}

We use the Alcoholic EEG dataset\footnote{https://kdd.ics.uci.edu/databases/eeg/eeg.data.html}
of 122 subjects divided into two groups: alcoholic and control groups.
Each subject completed about 100 trials and the data contains about
12K trials in total. For each trial, the subject was presented with
three different types of stimuli in $1$ second. EEG signals have
64 channels sampled at the rate of 256 Hz. Thus, each trial consists
of $64\times256$ samples in total. We convert the signals into spectrograms
using Short-Time Fourier Transform (STFT) with Hamming window of length
$64$ and $56$ overlapping samples. The signals were detrended by
removing mean values along the time axis. Because the signals are
real-valued, we only take half of the frequency bins. We also exclude
the first bin which corresponds to zero frequency. This results in
a tensor of shape $64\times32\times25$ where the dimensions are channel,
frequency and time, respectively. Fig.~\ref{fig:EEG_Spec} shows
examples of the input spectrogram of an alcoholic subject in 4 channels
which reveals some spatial correlations across channels and frequencies.
For this dataset, we randomly separate the trials of each subject
with a proportion of 0.6/0.2/0.2 for training/testing/validation.

To model the frequency change of all channels over time, we use LSTMs
\cite{hochreiter1997long}. We choose vector LSTMs with 200 hidden
units which show to work best in the experiment with Moving MNIST.
For matrix LSTMs, we select the one with hidden shape of $100\times100$.
Inputs to LSTMs are plain spectrogram which are sequences of matrices
of shape $64\times32$. For vector LSTM, these matrices are flattened
into vectors. 

As seen in Tab.~\ref{tab:EEG_result}, the vector LSTM with raw input
(Model 1) not only achieves the worse result but also consumes a very
large number of parameters. The matrix LSTM (Model 2) improves the
result by a large margin (67.7\% relative error reduction) while having
an order of magnitude fewer parameters than of Model 1. 

\begin{table}[H]
\begin{centering}
\begin{tabular}{|l|r|c|}
\hline 
\emph{Model} & \emph{\# Params} & \emph{Err (\%)}\tabularnewline
\hline 
\hline 
vec-LSTM & 1,844,201 & 5.29\tabularnewline
mat-LSTM & \textbf{160,601} & \textbf{1.71}\tabularnewline
\hline 
\end{tabular}
\par \end{centering}
\caption{Results on EEG classification.\label{tab:EEG_result}}
\end{table}

\subsection{Multi-Attention Matrix for Graph Modeling \label{subsec:Exp-Multiple-Attention}}

From existing work \cite{kipf2016semi}, we select 3 large citation
network datasets: Citeseer, Cora and PubMed, whose statistics are
reported in Tab.~\ref{tab:GraphDescription}. In each dataset, nodes
are publications and (undirected) edges are citation links between
those publications. Each node is represented as a bag-of-words feature
vector and is assigned with a label indicating the type of the publication.
The task is node classification. Similar to \cite{kipf2016semi},
we randomly select $1000$ nodes for testing and leave the remaining
for training/validation. The number of nodes for validation is set
to $1000$ for PubMed and $100$ for Cora and Citeseer.

\begin{table}
\begin{centering}
\begin{tabular}{|c|c|r|r|r|c|c|c|}
\hline 
\multirow{2}{*}{Dataset} & \multirow{2}{*}{\#Classes} & \multirow{2}{*}{\#Nodes} & \multirow{2}{*}{\#Edges} & \multirow{2}{*}{\#Features} & \multicolumn{3}{c|}{Node Degree}\tabularnewline
\cline{6-8} 
 &  &  &  &  & Max & Min & Avg\tabularnewline
\hline 
Citeseer & 6 & 3,312 & 4,732 & 3,703 & 99 & 1 & 2.78\tabularnewline
Cora & 7 & 2,708 & 5,429 & 1,433 & 168 & 1 & 3.90\tabularnewline
PubMed & 3 & 19,717 & 44,338 & 500 & 171 & 1 & 4.50\tabularnewline
\hline 
\end{tabular}
\par \end{centering}
\caption{Statistics of citation network datasets. \label{tab:GraphDescription}}
\end{table}

Our models used in this experiment are derived from the Column Networks
(CLN) \cite{pham2017column}. We implemented three variants. The default
variant uses mean pooling to aggregate information from neighbors,
as in Eq.~(\ref{eq:vec-CLN-aggregate}). For the other two variants,
vector CLN simply flattens the matrix of neighbors into vector and
multi-attention CLN applies multi-attention mechanism presented in
Sec.~\ref{subsec:Multiple-Attention-Column-Networ}.  

Our settings are similar for all three models: $50$ for the number
of neighbors per node (randomly sampled during training), $100$ for
the number of hidden units in case of the PubMed dataset and $20$
for the other 2 datasets, $5$ for the height of the column networks
(the range within which a node can receive messages from its neighbors)
and $5\times10^{-4}$ for L2 regularization. Besides, the number of
attention is set to $10$ in case of multi-attention CLN. Its means
that our attention matrix will have the shape $10\times50$. We train
the models using $100$ epochs. Early stopping is triggered if the
validation loss does not improve after 10 consecutive epochs. The
best test results of each model from $10$ different runs are reported
in Tab.~\ref{tab:GraphResults}. 

\begin{table}[H]
\begin{centering}
\begin{tabular}{|c|c|c|c|}
\hline 
Model & Citeseer & Cora & PubMed\tabularnewline
\hline 
\hline 
vector CLN & 70.2\% & 79.2\% & 88.8\%\tabularnewline
\hline 
CLN & 72.3\% & 81.4\% & 89.1\%\tabularnewline
\hline 
multi-attention CLN & \textbf{72.6\%} & \textbf{81.7\%} & \textbf{89.7\%}\tabularnewline
\hline 
\end{tabular}
\par \end{centering}
\caption{Results of node classification on citation networks.\label{tab:GraphResults}}
\end{table}

Under the same settings, vector CLN achieves the worst results in
all three datasets. It suggests that vectorizing the neighbor matrix
is not a good strategy in graph modeling, especially when the graph
is sparse, since most of the weights are wasteful. Meanwhile, aggregation
methods like the default CLN and multi-attention CLN can easily avoid
this problem and provide better results. We also observed that multi-attention
CLN slightly outperforms the default CLN. Our intuition is that multi-attention
CLN allows a node to accumulate more information. We also provide
the 3D t-SNE \cite{van2008visualizing} visualization of the last
hidden representations of all model on the test PubMed dataset for
further understanding. Interestingly, they exhibit 3 distinct patterns.
For the vector CLN, the points seem to contract around a line while
for the default CLN, they lie on a nearly perfect plane (although
we embed them on 3D). In case of the multi-attention CLN, we, instead,
observe a curved surface. We think this should be investigated more
in the future work.

\begin{figure}
\begin{centering}
\subfigure[vector CLN]{
	\includegraphics[height=0.18\textwidth]{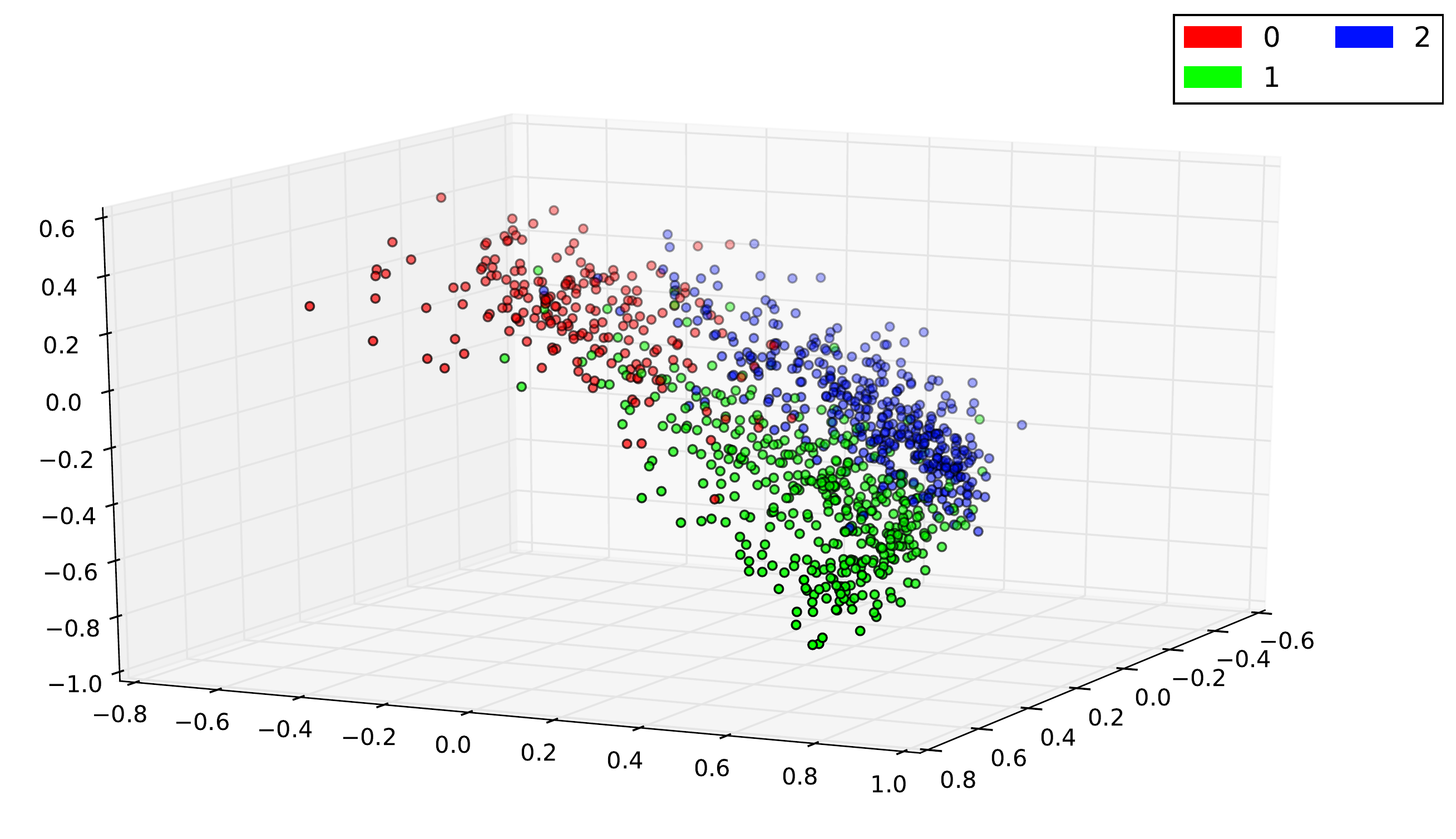}
}
\subfigure[CLN]{
	\includegraphics[height=0.18\textwidth]{default-CLN}
}
\subfigure[multi-attention CLN]{
	\includegraphics[height=0.18\textwidth]{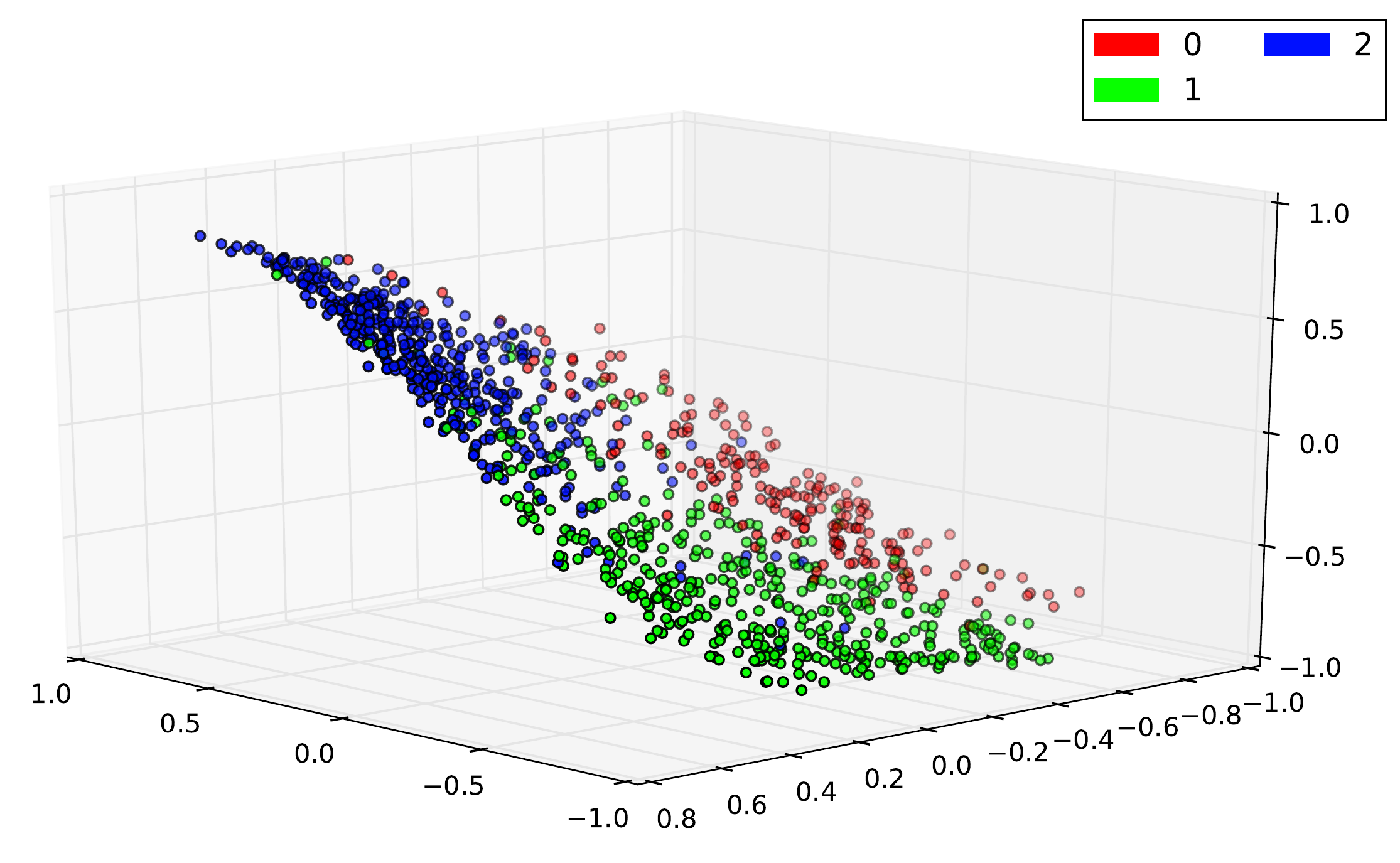}
}
\par \end{centering}
\caption{t-SNE visualization of the last hidden states in three variants of
CLN. Best viewed in color\label{fig:GraphEmbed}}
\end{figure}

\section{Related Work}

Matrix data modeling has been well studied in shallow or linear settings,
such as 2DPCA \cite{yang2004two}, 2DLDA \cite{ye2004two}, Matrix-variate
Factor Analysis \cite{xie2008matrix}, Tensor RBM \cite{nguyen2015tensor,qi2016matrix}.
Except 2DPCA and 2DLDA, all other methods are probabilistic models
which use matrix mapping to parameterize the conditional distribution
of the observed random variable given the latent variable. However,
since these models are shallow, their applications of matrix mapping
are limited. 

Deep learning inspired models for handling multidimensional data
including Multidimensional RNNs \cite{graves2009offline}, Grid LSTMs
\cite{kalchbrenner2015grid} and Convolutional LSTMs \cite{xingjian2015convolutional}.
The main idea of the Multidimensional RNNs and Grid LSTMs is that
any spatial dimension can be considered as a temporal dimension. They
extend the standard recurrent networks by making as many new recurrent
connections as the dimensionality of the data. These connections allow
the network to create a flexible internal representation of the surrounding
context. Although Multidimensional RNNs and Grid LSTMs are shown to
work well with many high dimensional datasets, they are complicate
to implement and have a very long recurrent loop (often equal to the
input tensor's shape) run sequentially. A convolutional LSTM, on the
other hand, works like a standard recurrent net except that its gates,
memory cells and hidden states are all 3D tensors with convolution
as the mapping operator. Consequently, each local region in the hidden
memory is attended and updated over time. However, not like our matrix
neural nets, applying convolutional LSTM to memories and graphs is
not straightforward. 

There has been a large amount of work on graph modeling recently.
Apart from those \cite{perozzi2014deepwalk,grover2016node2vec} based
on skip-gram model \cite{mikolov2013distributed} to learn node representations
(by using random walk to find node neighborhoods), other methods explicitly
exploit the graph structure (via its adjacency matrix) to exchange
states between nodes. Although they originate from different viewpoints,
e.g. spectral graph theory \cite{defferrard2016convolutional,kipf2016semi}
and message propagation \cite{scarselli2009graph,li2016gated,pham2017column},
their formulations (with some restrictions like no edge types) are,
indeed, special cases of our matrix neural nets (see Sec.~\ref{sec:Matrix-Rep-of-Graph}
and the Appx.~\ref{subsec:Graph-Convolution-as}).

\section{Discussion}

This study investigated an alternative distributed representation
in neural nets where information is distributed across neurons arranged
in a matrix. This departs from the existing canonical representation
using vectors. Comprehensive experimental results have demonstrated
that our matrix models perform significantly better than the vector
counterparts when data are inherently matrices. Besides, matrix representation
is naturally in line with recent memory-augmented RNNs and graph neural
networks. This suggest a new way of thinking and opens a wide room
for future work.

\section*{}

\appendix

\section{Matrix GRU \label{subsec:Matrix-GRU}}

Likewise, the GRU block is specified as:

\begin{eqnarray*}
Z_{t} & = & \text{sigm}(\mat_{2}(X_{t},H_{t-1};\thetab_{z}))\\
R_{t} & = & \text{sigm}(\mat_{2}(X_{t},H_{t-1};\thetab_{r}))\\
\tilde{H}_{t} & = & \sigma\left(\mat_{2}(X_{t},R_{t}\odot H_{t-1};\thetab_{h})\right)\\
H_{t} & = & (1-Z_{t})\odot H_{t-1}+Z_{t}\odot H_{t-1}
\end{eqnarray*}
where $R_{t}$ is the reset gate, and $Z_{t}$ is the interpolation
factor.

\section{Graph Convolution as Matrix Operation \label{subsec:Graph-Convolution-as}}

In this section, we demonstrate how Graph Convolutional Network (GCN)
\cite{kipf2016semi} - a model that leverages spectral graph theory
to efficiently apply convolution operator on graph can be seen as
a special case of our matrix feed-forward nets.

To begin, we will reuse the graph notation $\Graph=(\Vertices,\Edges)$
from Sec.~\ref{sec:Matrix-Rep-of-Graph}. $A$ is still the adjacency
matrix of $\Graph$. $D\in\Real^{|\Vertices|\times|\Vertices|}$ be
the diagonal \textit{degree matrix} satisfied that $D_{i,i}=\sum_{j}A_{i,j}$.
A normalized graph Laplacian $L\in\Real^{|\Vertices|\times|\Vertices|}$
is defined as $L=\mathrm{I}-D^{-\frac{1}{2}}AD^{-\frac{1}{2}}$ in
case A is symmetric and $L=\mathrm{I}-D^{-1}A$ in case $A$ is asymmetric;
$\mathrm{I}$ is the identity matrix of size $|\Vertices|\times|\Vertices|$. 

In spectral graph theory, the convolution operation is defined as
a product between a filter $g$ and a graph signal $\xb\in\Real^{|\Vertices|}$
(here, we only assume real value for each node) over the Fourier domain:

\begin{equation}
g\star\xb=Ug(\Lambda)U^{\intercal}\xb\label{eq:SpectralGraph-1}
\end{equation}
where $U\in\Real^{|\Vertices|\times|\Vertices|}$ is the matrix of
eigenvectors and $\Lambda\in\Real^{|\Vertices|\times|\Vertices|}$
is the diagonal matrix of eigenvalues in the Eigen Decomposition of
$L=U\Lambda U^{\intercal}$. In Eq.~(\ref{eq:SpectralGraph-1}),
$U^{\ensuremath{\intercal}}\xb$ can be seen as the Fourier transform
of $\xb$ while $U(g(\Lambda)U^{\intercal}\xb)$ is the inverse Fourier
transform of the convolutional result $g_{\ensuremath{\thetab}}(\Lambda)U^{\intercal}\xb$
to the spatial domain. In fact, computing Eq.~(\ref{eq:SpectralGraph-1})
directly are very expensive for large graphs which requires $O(|\Vertices|^{2})$
time complexity. Therefore, \cite{defferrard2016convolutional} propose
to replace the non-parametric function $g(\Lambda)$ with a polynomial
approximation using the truncated Chebyshev expansion of order $K-1$:

\begin{equation}
g(\Lambda)\approx g_{\thetab}(\Lambda)=\sum_{k=0}^{K-1}\theta_{k}T_{k}(\tilde{\Lambda})\label{eq:FilterApprox-1}
\end{equation}
where the parameter $\thetab=\{\theta_{0},...,\theta_{K-1}\}\in\Real^{K}$
is a vector of Chebyshev coefficients and $T_{k}(.)$ is the $k$-order
Chebyshev polynomial computed by the recursive relation $T_{k}(x)=2xT_{k-1}(x)-T_{k-2}(x)$
with $T_{\ensuremath{0}}=1$ and $T_{1}=x$. $\tilde{\Lambda}=2\Lambda/\lambda_{\text{max}}-\mathrm{I}_{|\Vertices|}$
is the scale version of $\Lambda$ to ensure that all the eigenvalues
lie in $[-1,1]$ ($\lambda_{\ensuremath{\text{max}}}$ denotes the
largest eigenvalue of $L$). Substitute $g(\Lambda)$ from Eq.~(\ref{eq:FilterApprox-1})
to Eq.~(\ref{eq:SpectralGraph-1}), we have:

\begin{eqnarray}
g\star\xb & \approx & Ug_{\thetab}(\tilde{\Lambda})U^{\intercal}\xb=U\left(\sum_{k=0}^{K-1}\theta_{k}T_{k}(\tilde{\Lambda})\right)U^{\intercal}\xb\nonumber \\
 & = & \left(\sum_{k=0}^{K-1}\theta_{k}T_{k}(U\tilde{\Lambda}U^{\intercal})\right)\xb=\left(\sum_{k=0}^{K-1}\theta_{k}T_{k}(\tilde{L})\right)\xb\label{eq:ChebyshevExpan-1}
\end{eqnarray}
where $\tilde{L}=2L/\lambda_{\text{max}}-\mathrm{I}$ is the scaled
Laplacian matrix. In stead of computing Eq.~(\ref{eq:ChebyshevExpan-1})
recurrently from $0$ to $K-1$, \cite{kipf2016semi} suggest building
a deep convolutional neural networks of $K$ layers and limit the
Chebyshev estimation at each layer to $1$-order (e.g. $K=1$). The
transformation at one layer of the GCNs then becomes:

\begin{eqnarray}
g\star\xb & \approx & \theta_{0}T_{0}(\tilde{L})\xb+\theta_{1}T_{1}(\tilde{L})\xb=\theta_{0}\xb+\theta_{1}\tilde{L}\xb\label{eq:GCN-1}
\end{eqnarray}
\cite{kipf2016semi} further assume that $\lambda_{\text{max}}\approx2$
to avoid eigenvalues decomposition of $L$, which is also expensive.
They believe that the neural network can adapt this change in scale
during training. Eq.~(\ref{eq:GCN-1}) is now equivalent to:

\begin{eqnarray*}
g\star\xb & \approx & \theta_{0}\xb+\theta_{1}(L-\mathrm{I})\xb=\theta_{0}\xb-\theta_{1}(D^{-\frac{1}{2}}AD^{-\frac{1}{2}})\xb
\end{eqnarray*}
Continue setting $\theta=\theta_{0}=-\theta_{1}$, they come up with:

\begin{eqnarray}
g\star\xb & \approx & \theta(\mathrm{I}+D^{-\frac{1}{2}}AD^{-\frac{1}{2}})\xb\approx\theta(\tilde{D}^{-\frac{1}{2}}\tilde{A}\tilde{D}^{-\frac{1}{2}})\xb\label{eq:GCN-2}
\end{eqnarray}
where $\tilde{A}=A+\mathrm{I}$ and $\tilde{D}_{i,i}=\sum_{j}\tilde{A}_{i,j}$.
$\mathrm{I}+D^{-\frac{1}{2}}AD^{-\frac{1}{2}}\approx\tilde{D}^{-\frac{1}{2}}\tilde{A}D^{-\frac{1}{2}}$
is the \textit{renormalization trick} to ensure numerical stability. 

Note that up to now, we still assume that each node is correspondent
with a real value. Actually, Eq.~(\ref{eq:GCN-2}) can be extended
to adapt the vector representation of nodes. In this case, we can
rewrite Eq.~(\ref{eq:GCN-2}) as:

\begin{equation}
Y=(\tilde{D}^{-\frac{1}{2}}\tilde{A}\tilde{D}^{-\frac{1}{2}})X\Theta\label{eq:GCN_matrix-1}
\end{equation}
Apparently, Eq.~(\ref{eq:GCN_matrix-1}) bears a resemblance to our
mat2mat formula (see Eq.~(\ref{eq:matrix-neuron})). The only difference
is that $\left(\tilde{D}^{-\frac{1}{2}}\tilde{A}\tilde{D}^{-\frac{1}{2}}\right)$
is not a parameter but a precomputed matrix.

{\small{}\bibliographystyle{plain}

}{\small \par}

\end{document}